\documentclass[conference]{IEEEtran}
\IEEEoverridecommandlockouts

\usepackage{balance}
\usepackage{cite}
\usepackage{amsmath,amssymb,amsfonts}
\usepackage{graphicx}
\usepackage{textcomp}
\usepackage{xcolor}
\usepackage{booktabs}
\usepackage{multirow}
\usepackage{array}
\usepackage{placeins}
\usepackage{stfloats}
\usepackage{subcaption}
\usepackage[hidelinks]{hyperref}

\newcolumntype{L}[1]{>{\raggedright\arraybackslash}p{#1}}

\begin{document}
\raggedbottom
\bstctlcite{IEEEexample:BSTcontrol}

\title{\textsc{Tide}: Trustworthy and Interpretable Battery Degradation Estimation with Contextual Learning and Symbolic Distillation}

\author{
\IEEEauthorblockN{
Wen Yang Tan\textsuperscript{1},
Jiawei Li\textsuperscript{1},
Fang Liu\textsuperscript{2},
Wei Zhang\textsuperscript{1,*},
Sumei Sun\textsuperscript{1},
Peng Cheng Wang\textsuperscript{1},
Elisa Y. M. Ang\textsuperscript{1}}
\IEEEauthorblockA{\textsuperscript{1}Singapore Institute of Technology (SIT), Singapore 828608}
\IEEEauthorblockA{\textsuperscript{2}Singapore University of Social Sciences (SUSS), Singapore 599494}
\IEEEauthorblockA{Emails: wenyang.tan@singaporetech.edu.sg, jiawei009@e.ntu.edu.sg, liufang@suss.edu.sg, wei.zhang@singaporetech.edu.sg,\\ sumei.sun@singaporetech.edu.sg, victor.wang@singaporetech.edu.sg,  elisa.ang@singaporetech.edu.sg}
\thanks{This research is supported by A*STAR under its MTC Individual Research Grants (IRG) (Award M23M6c0113), MTC Programmatic (Award M23L9b0052), and the Shanghai Sci-tech Co-research Program (Award 25HB2702600). (\textit{Corresponding author: Wei Zhang})}
}

\maketitle

\begin{abstract}
Battery health estimation is fundamental for battery management in battery-powered systems, where inaccurate health states may affect control, maintenance, and service life. It becomes even more critical in intelligent connected systems, where estimation errors can propagate across interconnected devices and downstream decisions. In this paper, we propose \textsc{Tide}, a \underline{t}rustworthy and \underline{i}nterpretable battery \underline{d}egradation \underline{e}stimator for reliable battery health estimation. \textsc{Tide} jointly considers accuracy, trustworthiness, and interpretability, which are all essential for practical deployment and downstream decision making. To realize these objectives, \textsc{Tide} combines battery-domain knowledge with operational measurements in a three-component backbone. A knowledge-guided degradation prior promotes trustworthy estimation, a monotone residual component provides interpretable aging-consistent refinement, and a contextual learning component captures battery-specific operational effects for improved accuracy. {\color{black}The trained backbone is then distilled into a compact symbolic surrogate to provide model-level interpretability and support deployment.} Experiments show that \textsc{Tide} achieves strong estimation accuracy, improving overall estimation fidelity by an average of 19.7\% over representative baselines. Its knowledge-guided prior and monotone residual modelling substantially reduce aging-consistency violations, supporting trustworthy estimation. Meanwhile, the backbone enables component-level interpretation, while symbolic distillation provides a compact model-level representation of the learned estimation logic. These results support the practical use of \textsc{Tide} for battery health monitoring and decision support in intelligent connected systems.
\end{abstract}


\section{introduction}
Rechargeable batteries have become a fundamental energy source for a wide range of battery-powered systems such as electric vehicles and renewable energy storage systems, and such systems are increasingly deployed in Web-of-Things (WoT) environments \cite{IEA25}. As their adoption continues to expand, understanding battery health has become increasingly important for ensuring safe, reliable, and efficient operation throughout the battery lifetime. Without reliable battery health assessment, batteries may experience accelerated degradation, unexpected failures, and even safety hazards. Battery state of health (SoH), commonly defined as the percentage of remaining capacity relative to the battery’s initial capacity, has become a standard indicator of battery health. For example, a new battery typically has an SoH of 100\%, while an SoH of 80\% is commonly used as an end-of-life (EoL) threshold. Modern battery management systems (BMSs) continuously monitor various SoH-related measurements and estimate battery SoH to support charging control, maintenance planning, and operational decision-making \cite{pradhan2022battery}.   

Considerable research efforts have been devoted to battery SoH, with the majority of existing studies formulating a battery health prognostics problem that predicts the future SoH of a battery from its historical degradation trajectory \cite{kwan5871665knowdiff}. Early studies mainly relied on traditional approaches such as equivalent circuit models \cite{berecibar2016critical}. As battery sensing data became increasingly available, data-driven machine learning (ML) and deep learning methods gradually became the dominant paradigm by directly learning degradation patterns from operational measurements \cite{wang2025dgat,bai2023convolutional,zhou2024graph,tao2024data}. More recently, physics-guided and knowledge-informed learning have further improved robustness by incorporating knowledge into the learning process \cite{pamshetti2026knowledge,navidi2024physics,wang2024physical}. To facilitate deployment in resource-constrained systems, lightweight and edge battery models have also developed with model compression and knowledge distillation \cite{dharshini2026smaller}. 

Despite these advances, battery SoH forecasting generally assumes that ground-truth SoH values are already available, for example from the BMS, as inputs for future prediction. However, accurate SoH calculation requires detailed capacity measurements over a complete charge or discharge cycle, from fully depleted to fully charged or vice versa. In practice, complete charging or discharging processes are uncommon. Even when such measurements are available, deriving accurate SoH remains challenging because the measured capacity can be affected by operating conditions and measurement uncertainties. Consequently, recent studies have begun to investigate battery SoH estimation or modelling, which infers the current SoH from BMS signals collected during normal battery operation \cite{liu2025data}. Promising estimation performance has been reported. Their broader suitability for practical battery systems, however, remains insufficiently examined.

Accuracy alone is insufficient. In real-world scenarios, degradation modelling should jointly consider accuracy, trustworthiness, and interpretability. While most existing studies remain accuracy-driven, trustworthiness requires estimated SoH trajectories to remain consistent with physically plausible aging behavior, whereas interpretability supports transparent verification and diagnosis. Early attempts have explored intrinsically interpretable models, including Kolmogorov–Arnold Networks (KANs), which learn explicit functional relationships that can be inspected and represented symbolically \cite{liu2025kan2}. However, directly applying KANs alone does not fully address SoH estimation, as accuracy may be compromised, the overall process remains difficult to interpret, and direct symbolic regression often produces overly complex expressions \cite{Cranmer2023PySR}. In this paper, we argue that no single information or model component is sufficient. Operational data capture battery-specific conditions but offer limited guidance on overall degradation, whereas degradation knowledge provides global aging constraints but cannot fully represent contextual variations. We combine these complementary measurement data with the knowledge model, and assign accuracy, trustworthiness, and interpretability to distinct ML components.

We propose \textsc{Tide}, a \underline{t}rustworthy and \underline{i}nterpretable battery \underline{d}egradation \underline{e}stimator for battery SoH estimation. Rather than relying on a single estimation model, \textsc{Tide} addresses different objectives through complementary components. It first integrates battery-domain knowledge with operational sensing measurements to construct a structured three-component backbone estimator. A knowledge-guided prior is introduced to provide a physically meaningful estimation foundation for accuracy and trustworthiness. Building upon this foundation, a monotone residual component refines the estimation while preserving the expected battery aging behavior. A contextual residual component then captures battery-specific operational variations that cannot be explained by the preceding components, further improving estimation accuracy. {\color{black}While this estimator backbone supports component-level interpretation, its model-level logic remains complex. For resource-constrained battery systems, repeated SoH estimation also motivates a compact representation that is simpler to evaluate. Here, we further apply symbolic distillation to approximate the trained backbone with an explicit mathematical expression. The symbolic surrogate supports model interpretation and offers an alternative for deployment without executing the backbone.}

Extensive experiments are conducted to evaluate \textsc{Tide}. We first compare \textsc{Tide} with representative SoH estimation methods in terms of estimation accuracy and trustworthiness, and then examine different \textsc{Tide} variants to assess the contribution of each component. The results show that \textsc{Tide} achieves high estimation fidelity, improving accuracy by an average of 19.7\% over the benchmarking methods. Beyond accuracy, \textsc{Tide} eliminates aging-consistency violations, whereas representative ML baselines exhibit noticeable violations. {\color{black}Its structured backbone and symbolic surrogate provide complementary interpretations and the computational analysis further examines the deployment implications of them.} Together, these results demonstrate that \textsc{Tide} offers an accurate, trustworthy, and interpretable solution for battery health estimation. More broadly, its design philosophy may inform future ML models for intelligent WoT systems that require reliable and transparent decision-making.

The rest of the paper is organized as follows. We present the technical details of \textsc{Tide}  in Section \ref{sec:method}. In Section \ref{sec:experiments}, we conduct an experimental study and present the experiment results. Finally, we conclude this paper in Section \ref{sec:conclusion}.

\section{\textsc{Tide} Framework} 
\label{sec:method}
In this section, we first formulate the SoH estimation problem and then present the proposed \textsc{Tide} framework. 

\subsection{Problem Formulation}

In practical battery systems, a BMS operates as a separate monitoring and control module that collects operational measurements, such as voltage, current, and temperature. These measurements provide partial observations of battery condition but do not directly reveal SoH. Battery SoH estimation is therefore formulated as a learning problem that infers the current health state from available operational features and, where applicable, complementary information like degradation knowledge. Formally, let $\mathbf{x}_i\in\mathbb{R}^{d}$ denote the input feature vector available at the $i$-th charging/discharging cycle or estimation instance. After optional preprocessing, the processed feature vector $\bar{\mathbf{x}}_i$ is fed into an SoH estimator $f(\cdot)$ to estimate the current SoH,
\begin{equation}
\hat{y}_i=f(\bar{\mathbf{x}}_i), \qquad \hat{y}_i\in[0,1],
\end{equation}
where $\hat{y}_i$ as a percentage denotes the predicted SoH. In the ML context, $f(\cdot)$ denotes the estimation model and focuses on estimation accuracy. Each estimation instance is processed independently, allowing the current SoH to be inferred from the available measurements and knowledge without requiring complete charge-discharge processes, historical data, or future observations.
Accuracy is the primary objective of most existing studies. However, practical battery health estimation also requires trustworthiness and interpretability. A model may achieve low estimation error while producing unreasonable SoH increases during battery aging, which undermines confidence in its estimates. Similarly, opaque estimation logic makes the results difficult to verify, diagnose, and deploy in practical battery management systems. Therefore, this work jointly targets accuracy, trustworthiness, and interpretability.

\subsection{Framework Overview}

\begin{figure}[!t]
    \centering
    \includegraphics[width=0.98\columnwidth]{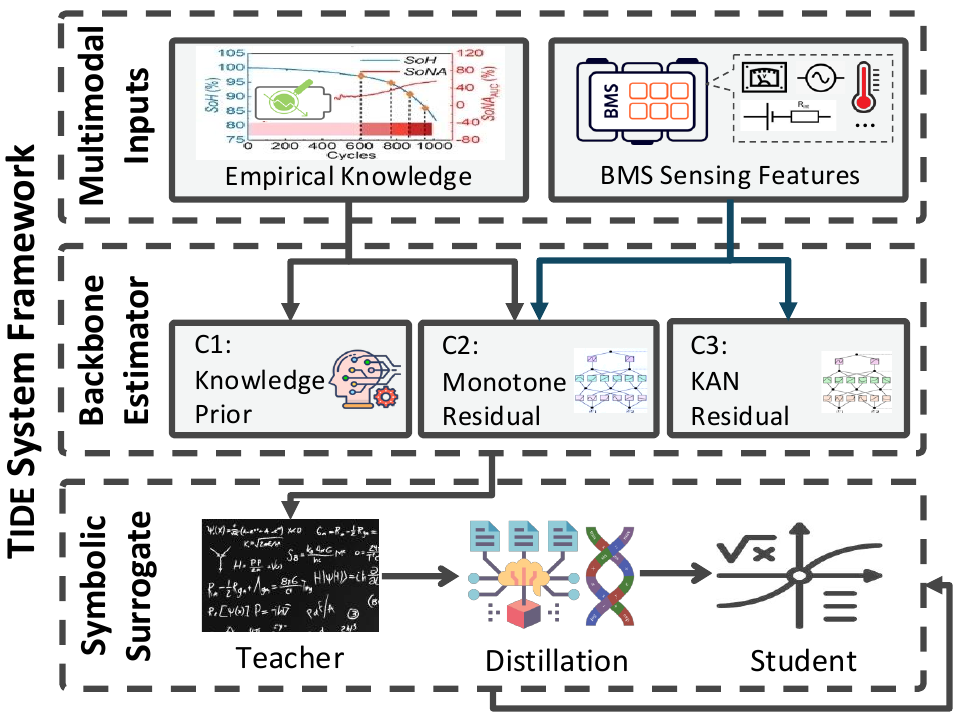}
    \caption{Overview of the proposed \textsc{Tide} framework. Degradation knowledge and BMS sensing features are jointly processed by a structured three-component backbone, {\color{black}followed by symbolic distillation to obtain a compact surrogate for model-level interpretation and deployment.}}
    \label{fig:tide-architecture}
\end{figure}

Fig. \ref{fig:tide-architecture} illustrates the overall architecture of the proposed \textsc{Tide} framework. Rather than directly mapping BMS sensing features to an ML model, \textsc{Tide} first extends the input space by introducing multimodal information sources. As shown in the first layer of the figure, empirical degradation knowledge provides global guidance on battery degradation patterns, while BMS sensing features characterize battery-specific operating conditions. 
While multimodal inputs provide richer information, they do not by themselves guarantee accurate, trustworthy, and interpretable estimation. To this end, we in the second layer introduce a structured backbone estimator consisting of three complementary components. The knowledge prior component $C_1$ first establishes a coarse estimation foundation from empirical battery knowledge, providing a physically meaningful starting point for estimation. Since this foundation alone cannot capture the fine-grained degradation process, the monotone residual component $C_2$ progressively refines the estimate while preserving the expected aging behavior. Component $C_3$ for contextual KAN residual further models the remaining battery-specific contextual effects to improve accuracy. Together, the components progressively construct a structured SoH estimator backbone with functional representations that facilitate component-level interpretation. {\color{black}The combined estimation logic nevertheless remains distributed across the backbone components.} 
{\color{black}The \textsc{Tide} backbone then serves as the teacher for symbolic distillation, which brings this logic into a single expression. The resulting symbolic surrogate approximates the backbone's SoH predictions, provides further interpretation, and facilitates \textsc{Tide} deployment.}

\subsection{\textsc{Tide} Backbone: High-Capacity Battery Health Estimator}
The backbone estimator is the core of the proposed \textsc{Tide} framework. This subsection presents its technical details.

\begingroup\color{black}
\subsubsection{Trustworthy Knowledge Prior}
\label{subsubsec:knowledge-prior-revised}

The first component, $C_1$, converts empirical battery-aging knowledge into a
coarse but physically meaningful SoH baseline. The prior knowledge used here is
obtained from established battery degradation studies and practical BMS
observations, i.e., battery SoH generally decreases as cycling accumulates, and
internal resistance tends to increase as electrochemical aging progresses
\cite{berecibar2016critical,pradhan2022battery,pamshetti2026knowledge,navidi2024physics,wang2024physical}. These two degradation indicators are also available from the deployment-safe BMS feature set, allowing the prior to be constructed without using capacity-derived or future information at inference time.
To incorporate this knowledge, we instantiate $C_1$ as an exponential-resistance
degradation baseline using the cycle index $i$ and internal resistance $r_i$,
\begin{equation}
C_1(i,r_i)=a\exp(-b i)-d r_i,
\label{eq:c1}
\end{equation}
where $a$, $b$, and $d$ are fitted using only the training batteries. The
exponential term models the smooth long-term decline associated with cycle
accumulation, and the resistance term reflects the negative association
between increasing internal resistance and battery health. Therefore, $C_1$ is
not introduced as a complete electrochemical model or as an additional black-box
learner. Instead, it serves as a transparent degradation anchor that initializes
the estimation around an aging-consistent trend. The remaining unexplained part
of the SoH target is then delegated to the residual components,
\begin{equation}
\delta_i=y_i-C_1(i,r_i),
\label{eq:residual}
\end{equation}
where $C_2$ learns the degradation-related monotonic correction and $C_3$ learns
the contextual correction from operational measurements. This design separates
the global aging prior from data-driven residual learning, so that the full
backbone can preserve physically meaningful degradation guidance while still
capturing battery-specific operating effects.
\endgroup

\subsubsection{Interpretable Monotone Residual}

The residual estimation error left by $C_1$ consists of two parts. The first is associated with degradation-related characteristics that is expected to be monotonic and follow the battery aging trend. The second captures battery-specific operational and contextual effects that do not necessarily exhibit monotonic behavior. Here, the monotone residual component $C_2$ is designed to model the former part using an empirically selected subset of degradation-related features such as cycle index and resistance. With these features as the input, a monotonic mapping can be imposed to ensure that the component remains consistent with the expected aging behavior.
Formally, the component is,
\begin{equation}
C_2(\tilde{\mathbf{x}}^{\mathrm{mon}}_i) = - \sum_{p=1}^{P}\sum_{h=1}^{H}\alpha_{ph}\,
\operatorname{softplus}\left(\beta_{ph}\tilde{x}^{\mathrm{mon}}_{ip}+\gamma_{ph}\right)+\beta_0,
\label{eq:monotone-component}
\end{equation}
where $\mathbf{x}^{\mathrm{mon}}_i\subset\mathbf{x}_i$ denotes the selected features and $\tilde{\mathbf{x}}^{\mathrm{mon}}_i$ is the corresponding standardized feature vector. Here, $P$ is the number of selected features and $H$ is the number of hidden units associated with each feature. The $\operatorname{softplus}(\cdot)$ function is adopted as a smooth non-decreasing activation. The effective weights are constrained to be non-negative by $\alpha_{ph}=\operatorname{softplus}(\tilde{\alpha}_{ph})\ge0$ and $\beta_{ph}=\operatorname{softplus}(\tilde{\beta}_{ph})\ge0$, where $\tilde{\alpha}_{ph}$ and $\tilde{\beta}_{ph}$ are unconstrained learnable parameters. The parameters $\alpha_{ph}$ and $\beta_{ph}$ control the contribution and sensitivity of each hidden unit, respectively, while $\gamma_{ph}$ and $\beta_0$ denote the activation shift and component bias. Since $\operatorname{softplus}(\cdot)$ is non-decreasing, the effective weights are non-negative, and Eq. (\ref{eq:monotone-component}) adopts a leading negative sign, so $C_2$ is guaranteed to be non-increasing with respect to each selected feature. 
Consequently, $C_2$ learns a degradation-aware residual that refines the coarse prior while preserving the expected monotonic aging behavior. Its explicit functional formulation also provides a transparent representation of the learned degradation relationship.

\subsubsection{Accurate and Interpretable KAN Residual}

The second part of the residual is associated with the battery-specific contextual information that does not necessarily follow the expected aging trend. Therefore, the third component, $C_3$, is designed to model this contextual residual using the remaining features. No monotonic constraint is imposed on $C_3$, allowing it to flexibly capture complex nonlinear operational characteristics. 
Formally, let $\mathbf{x}_i^{\mathrm{ctx}}=\mathbf{x}_i\setminus\mathbf{x}_i^{\mathrm{mon}}$
denote the remaining features after removing the monotonic degradation features used by $C_2$. The contextual residual component is implemented using the latest MultKAN architecture~\cite{liu2025kan2}. Compared with conventional ML algorithms, MultKAN provides greater flexibility in modeling complex nonlinear contextual effects while preserving an explicit functional representation. Overall, $C_3$ is introduced to provide a flexible nonlinear residual correction that further improves estimation accuracy. 
\subsubsection{Backbone Integration and Learning}

The proposed backbone integrates the three complementary components into a unified end-to-end learning framework. Specifically, $C_1$ establishes a coarse degradation prior, $C_2$ learns a knowledge-guided monotonic correction, and $C_3$ captures the remaining contextual correction. Given residual
$\delta_i$, the correction is standardized as $\tilde{\delta}_i=T(\delta_i)$ to improve optimization stability. Thus, the final SoH estimation is then obtained by,
\begin{equation}
\hat{y}_i=C_1(i,r_i)+T^{-1}\big(C_2(\tilde{\mathbf{x}}^{\mathrm{mon}}_i)+C_3(\tilde{\mathbf{x}}^{\mathrm{ctx}}_i)\big),
\label{eq:tide-backbone}
\end{equation}
where $T^{-1}(\cdot)$ denotes the inverse standardization transform that maps the predicted degradation correction back to the original SoH scale. All three components are jointly trained in an end-to-end manner, allowing the backbone to balance degradation consistency with contextual flexibility.
The backbone is trained by minimizing the loss function,
\begin{equation}
\mathcal{L} = \mathcal{L}_{\mathrm{pred}} 
+ 
\lambda_{\mathrm{cyc}}\mathcal{L}_{\mathrm{cyc}}
+
\lambda_{\mathrm{ir}}\mathcal{L}_{\mathrm{ir}}
+
\lambda_{\mathrm{range}}\mathcal{L}_{\mathrm{range}}
+
\lambda_{\mathrm{aux}}\mathcal{L}_{\mathrm{aux}},
\label{eq:training-loss}
\end{equation}
where $\mathcal{L}_{\mathrm{pred}}$ minimizes the prediction error of the standardized degradation correction, $\mathcal{L}_{\mathrm{cyc}}$ and $\mathcal{L}_{\mathrm{ir}}$ enforce degradation consistency with respect to feature cycle index and internal resistance, respectively, $\mathcal{L}_{\mathrm{range}}$ constrains the predicted SoH within the feasible operating range, and $\mathcal{L}_{\mathrm{aux}}$ regularizes the magnitude of the learned correction. The weighting coefficients $\lambda_{\mathrm{cyc}}$, $\lambda_{\mathrm{ir}}$, $\lambda_{\mathrm{range}}$, and $\lambda_{\mathrm{aux}}$ balance the corresponding objectives during joint training.
This unified design enables the proposed estimator to simultaneously achieve accurate, trustworthy, and interpretable battery SoH estimation.

\subsection{\textsc{Tide} Symbolic Surrogate}

{\color{black}The symbolic surrogate extends the component-level explanations of the backbone to the complete SoH estimate. Its compact expression makes the learned input--output relationship easier to inspect and provides a candidate for direct evaluation in resource-constrained battery systems.}
{\color{black}The trained \textsc{Tide} backbone serves as the teacher, and symbolic distillation learns a surrogate satisfying $S(\mathbf{x}) \approx f_{\textsc{Tide}}(\mathbf{x})$, where $f_{\textsc{Tide}}(\cdot)$ denotes the backbone and $S(\cdot)$ denotes the symbolic expression. Each estimator uses its corresponding fitted input preprocessing. Because the target is the complete SoH prediction, evaluating the surrogate does not require separate execution of the backbone components.}
{\color{black}
Symbolic search uses cycle index, internal resistance, its five-cycle
rolling mean, mean voltage, voltage variability, log-transformed cycle
index, and the cycle--resistance interaction. These features combine
degradation indicators with operating context. Median imputation and
standardization are fitted on the sampled training records and applied
unchanged to validation and test inputs.
Specifically, we use PySR~\cite{Cranmer2023PySR} to approximate the backbone's SoH predictions
using addition, subtraction, multiplication, and square roots, subject
to a maximum expression size. The candidate with the lowest RMSE against
measured validation SoH is selected. Teacher-fidelity error is recorded
separately to distinguish agreement with the backbone from estimation
accuracy. The selected expression uses the same fitted preprocessing
at inference.
}

\begingroup\color{black}
\subsection{Efficiency Considerations}
\label{subsec:efficiency-considerations}

{\color{black}The proposed \textsc{Tide} framework separates offline model construction from repeated SoH inference. Backbone training and symbolic search can be performed offline, while the selected estimator is evaluated using the operational features available at deployment. In addition to providing model-level interpretation, the symbolic surrogate offers a compact alternative to the backbone. Its explicit expression can be evaluated using the fitted preprocessing without executing the neural components. This motivates its use in battery systems where model storage and computational resources are limited.
The benefit of this compact representation must be considered alongside its approximation accuracy. The expression's coefficients, preprocessing, and intermediate calculations still require resources, so expression compactness alone does not establish the memory or latency of a deployed implementation. The efficiency analysis in the following section examines parameter storage and inference cost alongside offline training cost, and the symbolic evaluation assesses SoH estimation error and teacher fidelity. Together, these analyses inform the choice between the backbone and its surrogate according to the application's accuracy and resource requirements.}
\endgroup

\section{Experiments}
\label{sec:experiments}
This section presents a comprehensive evaluation of the proposed \textsc{Tide} framework. We first describe the experimental setup, followed by quantitative and qualitative analyses to assess its estimation performance. 

\subsection{Experimental Setup}

Experiments are conducted on the public MIT-Stanford fast-charging lithium-ion battery cycle-life dataset \cite{severson2019data}, which contains operational measurements collected throughout the battery aging process. Following feature engineering and data cleaning, the dataset contains 140 batteries with 114,738 cycle-level records. The battery SoH is only used for performance evaluation and is computed as the ratio between the current discharge capacity and the initial rated capacity. To evaluate generalization to previously unseen batteries, a battery-disjoint split of 70\%, 15\%, and 15\% is adopted for training, validation, and testing, respectively. 
A total of 34 deployment-safe operational sensing features are extracted from the BMS, including cycle index, internal resistance, charging policy, temperature, voltage/current statistics, and derived interactions. All target-derived variables, such as SoH, degradation indicators, remaining useful life, capacity-retention ratios, and future battery information, are excluded to prevent information leakage.

The proposed \textsc{Tide} framework is implemented in Python using PyTorch, pykan, and PySR. The backbone is trained end-to-end using the \texttt{Adam} optimizer with an initial learning rate of $0.001$, a batch size of $2,048$, and early stopping based on the validation loss. The contextual KAN residual component employs the standard MultKAN architecture with a hidden width of $24$, a grid size of $3$, and cubic B-spline basis functions. All experiments are conducted on a Windows 11 workstation equipped with an Intel x64 CPU, $16$GB RAM, and an NVIDIA GeForce MX450 GPU with $2$GB memory. 
Performance is evaluated using Root Mean Squared Error (RMSE) and the coefficient of determination ($R^2$). Both metrics are computed separately for each of the test batteries and reported as average values. The standard deviation (STD) of the per-battery RMSE is additionally reported to evaluate estimation consistency and robustness across unseen batteries.

\subsection{Results and Analysis}
{\color{black}In this part, we evaluate the \textsc{Tide} framework from complementary perspectives, such as accuracy, trustworthiness, efficiency, and interpretability.} 

\begin{table}[!t]
\caption{SoH estimation performance of the compared algorithms. Lower RMSE and STD and higher $R^2$ indicate better performance. \textsc{Tide}’s improvement over each baseline is reported as the percentage increase in $R^2$.}
\label{tab:main-results}
\centering
\renewcommand{\arraystretch}{1.2}
\begin{tabular}{@{}cccccc@{}}
\toprule
Category & Model & RMSE & $R^2$ & STD & Improv. \\
\midrule
$-$
& \textsc{Tide} (ours) & 0.0081 & \textbf{0.964} & \textbf{0.0047} & $-$ \\
\midrule
\multirow{2}{*}{\shortstack{Hybrid}}
& CNN-BiGRU~\cite{liu2025data}
& \textbf{0.0080} & 0.955 & 0.0056 & 0.94\\
& CNN-BiLSTM~\cite{lien2024hybrid}
& 0.0085 & 0.950 & 0.0074 & 1.47\\
\midrule
\multirow{3}{*}{\shortstack{Sequence}}
& GRU~& 0.0136 & 0.867 & 0.0118 & 11.19\\
& LSTM~& 0.0129 & 0.845 & 0.0136 & 14.08\\
& TCN~& 0.0166 & 0.792 & 0.0132 & 21.72\\
\midrule
KAN
& Vanilla KAN~\cite{liu2025kan}
& 0.0318 & 0.572 & 0.0113 & 68.53\\
\bottomrule
\end{tabular}
\end{table}

\subsubsection{\textsc{Tide} Backbone SoH Estimation Accuracy}
We first evaluate the SoH estimation accuracy of the \textsc{Tide} backbone. The comparison includes representative state-of-the-art SoH estimation methods and forecasting models reconfigured for the same-cycle estimation task.
Table \ref{tab:main-results} shows that \textsc{Tide} achieves highly competitive performance while providing the best overall goodness-of-fit and estimation consistency. Although CNN-BiGRU attains a marginally lower RMSE ($0.0080$ vs. $0.0081$), \textsc{Tide} achieves the highest $R^2$ of $0.964$ together with the lowest STD of $0.0047$, corresponding to a $0.94\%$ improvement in $R^2$ over the strongest competing method. These results indicate that the proposed knowledge-guided backbone captures battery degradation characteristics more comprehensively and generalizes more consistently across different batteries.

Among the baselines, the two hybrid models, CNN-BiGRU and CNN-BiLSTM, substantially outperform the conventional sequence models, confirming the benefit of combining convolutional feature extraction with temporal learning. In contrast, the vanilla KAN baseline achieves an $R^2$ of only $0.572$, despite its benefit of explicit functional representations. This suggests that KAN alone does not provide accurate estimation without incorporating degradation knowledge. Overall, the \textsc{Tide} backbone demonstrates that improving trustworthiness and interpretability does not compromise estimation accuracy.

\begin{table}[!t]
\caption{Ablation study of \textsc{Tide} and its variants. Lower RMSE and STD and higher $R^2$ indicate better performance. \textsc{Tide}'s improvement over each variant is reported as the percentage increase in $R^2$.}
\label{tab:module-ablation}
\centering
\renewcommand{\arraystretch}{1.2}
\begin{tabular}{@{}cccccc@{}}
\toprule
Category & Model & RMSE & $R^2$ & STD & Improv. \\
\midrule
$-$ & \textsc{Tide} (ours) & \textbf{0.0081} & \textbf{0.964} & \textbf{0.0047} & $-$\\\midrule
\multirow{3}{*}{\shortstack{Variant}} & $C_1$ & 0.0397 & 0.318 & 0.0133 & 203.1\\
& $C_1+C_2$ & 0.0356 & 0.194 & 0.0179 & 396.9\\
& $C_1+C_3$ & 0.0136 & 0.896 & 0.0089 & 7.59\\
\bottomrule
\end{tabular}
\end{table}

{\color{black}
Table~\ref{tab:module-ablation} further examines the role and contribution of the prior
knowledge in the proposed backbone. The prior alone, $C_1$, achieves an $R^2$ of
0.318, indicating that empirical degradation knowledge provides a reasonable but coarse estimation foundation. This result also shows that the
prior by itself is not sufficient for accurate SoH estimation, because real
battery behavior is affected by operating conditions that cannot be fully
captured by cycle index and internal resistance alone. Adding the contextual KAN
residual component improves the model from $C_1$ to $C_1+C_3$, increasing $R^2$
from 0.318 to 0.896 and reducing RMSE from 0.0397 to 0.0136. However,
Table~\ref{tab:physical-consistency} shows that $C_1+C_3$ still produces an
internal-resistance MVR of 3.01\%, indicating that accuracy-driven contextual
learning can introduce local aging-inconsistent behavior when it is not paired with a monotonic correction. The full \textsc{Tide} backbone combines the prior,
monotone residual, and contextual residual components, achieving the strongest
reported ablation performance with RMSE 0.0081, $R^2=0.964$, and minimal MVR. These results support the intended role of knowledge, which provides an aging-consistent foundation, where the two
residual branches contribute complementary benefits
rather than simply forming an ensemble.}

More importantly, integrating both residual components further improves the $R^2$ from $0.896$ to $0.964$ while reducing the estimation STD from $0.0089$ to $0.0047$. Although the additional $R^2$ improvement is only $7.59\%$, it confirms that the monotone residual component complements rather than competes with the contextual component. By combining trustworthy degradation modelling with flexible contextual learning, the complete \textsc{Tide} backbone with all three components achieves the best overall estimation performance.

\begin{figure}[!t]
\centering
\begin{subfigure}[t]{\columnwidth}
    \centering
    \includegraphics[width=0.9\linewidth]{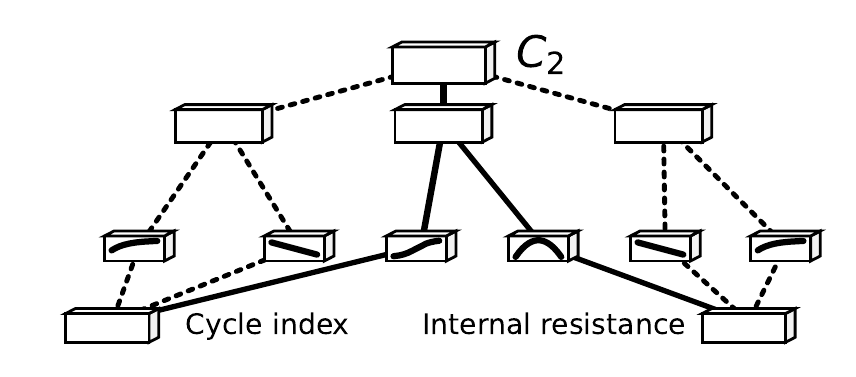}
    \caption{Monotone Residual Component $C_2$}
    \label{fig:tree-c2}
\end{subfigure}
\hfill
\begin{subfigure}[t]{\columnwidth}
    \centering
    \includegraphics[width=0.98\linewidth]{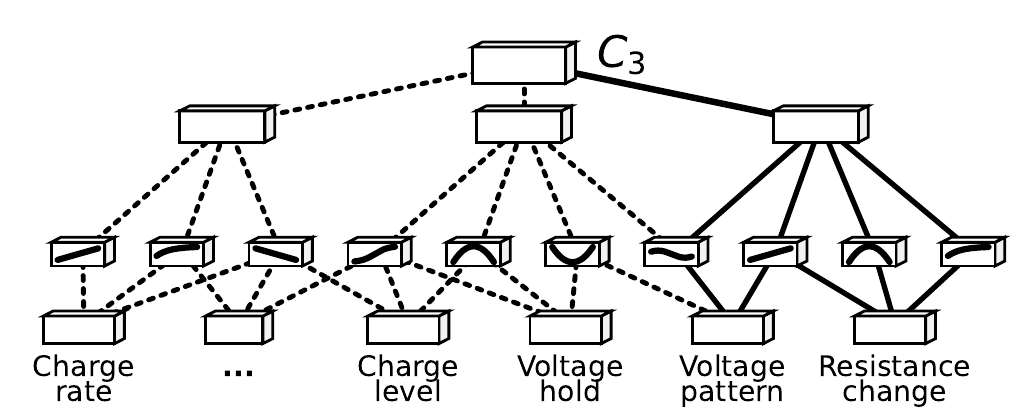}
    \caption{Contextual Residual Component $C_3$}
    \label{fig:tree-c3}
\end{subfigure}
\caption{Learned activation structures of the \textsc{Tide} backbone. Curve blocks denote learned KAN edge functions. Fig. \ref{fig:tree-c2}: $C_2$ combines cycle index and internal resistance for aging-related correction. Fig. \ref{fig:tree-c3}: $C_3$ models multiple operational features. Solid (dashed) paths indicate stronger (weaker) contributions.}
\label{fig:tree}
\end{figure}

\subsubsection{Trustworthy Aging-Consistent Behavior}
Accurate estimation alone is insufficient for practical battery health monitoring, and models that are accurate may not be trustworthy enough. In this part, we evaluate the trustworthiness of \textsc{Tide} and comparison models.
Trustworthiness can be evaluated from multiple aspects. Here, we adopt aging consistency as a representative diagnostic based on well-established battery degradation knowledge. 
Specifically, the cycle index $i$ and internal resistance $r_i$ are  two of the most widely recognized indicators of battery aging and investigated here. In general, battery health is expected to decline as either quantity increases. So, we evaluate the Monotonic Violation Rate (MVR), which measures the proportion of perturbations for which the predicted SoH increases when either $i$ or $r_i$ is increased while the remaining input features are held fixed. Lower MVR values indicate better adherence to the expected degradation behavior, and an MVR of $0\%$ indicates no observed monotonicity violation under this representative evaluation protocol.

{\color{black}
For each test instance, cycle index and internal resistance are
perturbed separately by adding $0.25$ in standardized feature space,
while all other inputs remain fixed. For sequence models, the
perturbation is applied only at the final timestep. The complete
estimator is reevaluated after each increase in cycle index or internal
resistance. A violation is counted if the same model predicts a higher
SoH after that increase than before it. Each
feature-specific MVR is the percentage of violations across all
evaluated test instances; the average reported in
Table~\ref{tab:physical-consistency} is the arithmetic mean of the
two rates. This diagnostic measures conditional model responses
to individual input changes, rather than the evolution of all
coupled battery measurements along a physical aging trajectory.
}

\begin{table}[!t]
\caption{Trustworthiness evaluation of \textsc{Tide}, comparison models, and \textsc{Tide} variants using MVR for cycle index $i$ and resistance $r_i$. Lower MVR values indicate better adherence to the expected battery aging behavior. }
\label{tab:physical-consistency}
\centering
\renewcommand{\arraystretch}{1.2}
\begin{tabular}{@{}ccccc@{}}
\toprule
Category & Model & $i$ MVR (\%) & $r_i$ MVR (\%) & Average (\%) \\
\midrule
$-$ 
& \textsc{Tide} (ours) & \textbf{0.00} & \textbf{0.00} & \textbf{0.00} \\
\midrule
\multirow{2}{*}{Hybrid}
& CNN-BiGRU & 31.48 & 91.90 & 61.69 \\
& CNN-BiLSTM & 27.34 & 36.53 & 31.94 \\
\midrule
KAN
& Vanilla KAN & 0.25 & 2.80 & 1.53 \\
\midrule
\multirow{3}{*}{Variant}
& $C_1$ & 0.0 & 0.0 & 0.0 \\
& $C_1+C_2$ & 0.0 & 0.0 & 0.0 \\
& $C_1+C_3$ & 0.0 & 3.01 & 1.51 \\
\bottomrule
\end{tabular}
\end{table}

Table~\ref{tab:physical-consistency} presents the trustworthiness evaluation of \textsc{Tide} and comparison models. \textsc{Tide} achieves zero MVR for both $i$ and $r_i$, resulting in an overall MVR of $0.0\%$. In contrast, although hybrid models achieve competitive estimation accuracy, they exhibit overall MVR values of $61.69\%$ and $31.94\%$, respectively. This indicates that conventional ML models may learn highly accurate mappings from data while still violating the expected battery degradation behavior, as no explicit aging-consistency constraints are imposed during learning. The vanilla KAN baseline achieves a much lower MVR of $1.53\%$, suggesting that its explicit functional representation naturally preserves certain monotonic characteristics. However, its estimation accuracy is much lower than \textsc{Tide}, demonstrating that interpretable model structures alone are insufficient for both accuracy and trustworthy.

The ablation study further reveals the contribution of each backbone component to trustworthy behavior. Both $C_1$ and $C_1+C_2$ maintain zero MVR, confirming that the trustworthy knowledge prior and the monotone residual component consistently preserve the expected aging direction. When the contextual KAN residual component $C_3$ is used without the monotone component $C_2$, the overall MVR increases to $1.51\%$, indicating that unconstrained contextual modelling can introduce violations despite its modelling flexibility. By integrating all components, \textsc{Tide} restores the MVR to $0.0\%$. Together, the components achieve trustworthy aging-consistent behavior while retaining the strong estimation capability.

\begingroup\color{black}

\begin{table}[!t]
\caption{Computational efficiency of the compared models. Parameter counts denote the total number of scalar entries in the neural model parameters. Memory denotes estimated FP32 parameter storage. As for the symbolic surrogate, 0 denotes the absence of neural parameters.}
\label{tab:efficiency-analysis}
\centering
\renewcommand{\arraystretch}{1.2}
\begin{tabular}{@{}crrrr@{}}
\toprule
\multirow{2}{*}{Model} & \multirow{2}{*}{\shortstack{\# of\\Parameters}} & \multicolumn{2}{c}{Time} & \multirow{2}{*}{\shortstack{Memory\\(KB)}} \\  & & Training (s) & Inference (ms) & \\
\midrule
LSTM & 9,793 & 111.38 & 0.0037 & 38.25 \\
GRU & 7,617 & 148.66 & 0.0067 & 29.75 \\
TCN & 8,609 & 127.01 & 0.0082 & 33.63 \\
\midrule
CNN-BiGRU & 21,185 & 331.09 & 0.0177 & 82.75 \\
CNN-BiLSTM & 25,409 & 224.19 & 0.0114 & 99.25 \\
\midrule
Vanilla KAN & 4,376 & \textbf{21.42} & 0.0120 & 17.09 \\
\midrule
\textsc{Tide}-Teacher & 11,797 & \multirow{2}{*}{1256.43} & 0.0247 & 46.08 \\
\textsc{Tide}-Symbolic & \textbf{0} & & \textbf{0.000005} & \textbf{0.00} \\
\bottomrule
\end{tabular}
\end{table}

\subsubsection{Efficiency Analysis}
\label{subsubsec:efficiency-analysis}

We further evaluate the computational efficiency of \textsc{Tide}. This analysis complements
the accuracy and trustworthiness results because a battery health estimator must
remain practical for repeated use in the BMS. Table~\ref{tab:efficiency-analysis} compares the computational efficiency of \textsc{Tide} with the representative baselines. The \textsc{Tide} teacher contains only 11.8K parameters, which is close to the individual ML models and approximately half the parameters required by the hybrid models. A similar pattern is observed for FP32 parameter memory footprint. The teacher requires a higher training cost, but this cost occurs offline and does not affect subsequent SoH estimation. Its inference latency is only less than 0.1 ms per instance, and the latency would generally be small enough for BMS health estimation, as SoH evolves slowly typically and meaningful changes only occur over multiple cycles. The symbolic surrogate further reduces the computational requirements, e.g., no neural network parameters, minimal memory footprint, and near-instant inference. The results demonstrate the practical efficiency of the \textsc{Tide} teacher and the additional lightweight advantage offered by its symbolic surrogate.
\endgroup

\begin{figure*}[!t]
\def \htmp{0.65\linewidth}
\centering
\begin{subfigure}[t]{0.5\columnwidth}
    \centering
    \includegraphics[height=\htmp]{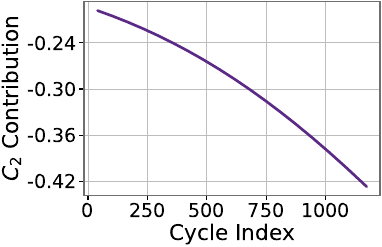}
    \caption{Cycle Index}
    \label{fig:feature-c}
\end{subfigure}
\begin{subfigure}[t]{0.5\columnwidth}
    \centering
    \includegraphics[height=\htmp]{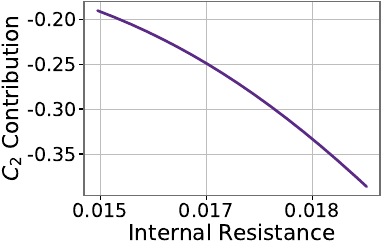}
    \caption{Internal Resistance}
    \label{fig:feature-r}
\end{subfigure}
\begin{subfigure}[t]{0.5\columnwidth}
    \centering
    \includegraphics[height=\htmp]{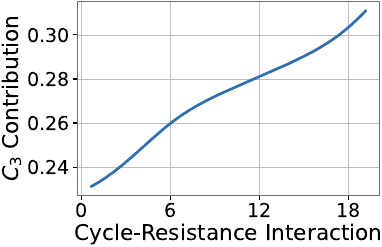}
    \caption{Cycle-resistance Interaction}
    \label{fig:feature-cr}
\end{subfigure}
\begin{subfigure}[t]{0.5\columnwidth}
    \centering
    \includegraphics[height=\htmp]{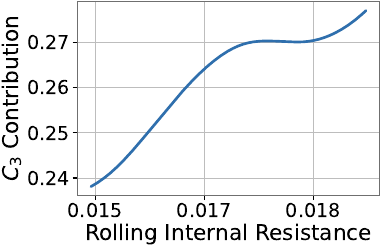}
    \caption{Rolling Internal Resistance}
    \label{fig:feature-roll}
\end{subfigure}
\caption{Representative learned response curves of \textsc{Tide}. Figs. \ref{fig:feature-c} and \ref{fig:feature-r} illustrate the monotone degradation responses learned by the degradation-aware component $C_2$, while Figs. \ref{fig:feature-cr} and \ref{fig:feature-roll} illustrate representative contextual responses learned by the contextual component $C_3$. The vertical axis denotes the contribution of each component to the SoH correction.}
\label{fig:feature}
\end{figure*}

\begingroup
\color{black}
\subsubsection{Interpretability of the Backbone}
\label{subsubsec:interpretability-evaluation}

{\color{black}The ablation study assesses the contribution of each backbone component, and the aging-consistency analysis evaluates the model's behavior under degradation-related changes. Here we examine how the components produce these adjustments. The analytical prior $C_1$ in Eq.~\eqref{eq:c1} provides an explicit degradation baseline, and the learned structures and response curves of $C_2$ and $C_3$ show how degradation-related and contextual features refine it. These component-level explanations also establish the basis for examining the complete estimator through symbolic distillation to be described later.}
Fig. \ref{fig:tree} presents the learned structures of the two residual components. The contextual KAN component $C_3$ is visualized using the KAN visualization toolkit~\cite{liu2025kan}, and the monotone component $C_2$ is represented using the same visual convention to show its input transformations and their aggregation. In Fig.~\ref{fig:tree-c2}, the stronger pathways combine cycle index and internal resistance, suggesting that the learned degradation correction depends on their joint contribution. In Fig.~\ref{fig:tree-c3}, prominent pathways involve resistance features and their interactions, with other features contributing through weaker pathways.
These structures make the distinct information processed by the two components visible, complementing the numerical results in
Table~\ref{tab:module-ablation}.

The response curves in Fig.~\ref{fig:feature} clarify how these transformations affect the residual correction. Each curve varies one feature while holding the remaining inputs at reference operating values. The vertical axis shows the corresponding component contribution to the residual in Eq.~\eqref{eq:residual}; negative and positive values respectively decrease and increase the estimate relative to the prior. As such, these curves describe conditional model responses, rather than causal effects of individual measurements on battery health.

In Figs.~\ref{fig:feature-c} and~\ref{fig:feature-r}, the contribution of $C_2$ becomes more negative as cycle index and internal resistance increase. The steeper responses at higher values indicate that the component applies stronger degradation corrections in those regions. This behavior is consistent with the intended role of the monotone component and complements the aging-consistency diagnostics in Table~\ref{tab:physical-consistency}. The positive corrections shown for $C_3$ in Figs.~\ref{fig:feature-cr} and~\ref{fig:feature-roll} illustrate how operating context can offset the baseline estimate under the plotted reference conditions. The nonlinear cycle--resistance response captures their combined influence, while the rolling-resistance response indicates that the correction varies with recent resistance conditions.

Together, the structural views and response curves explain how the backbone combines a degradation prior with learned residual adjustments. They make individual component responses inspectable, although the complete estimate still depends on their interaction. 
\endgroup

\begin{table}[!t]
\caption{Distillation performance from different teacher models to symbolic surrogates. Teacher, surrogate, and their gap RMSE measure SoH estimation error, while fidelity RMSE measures the approximation error between each surrogate and its teacher. Lower values indicate better performance.}
\label{tab:surrogate}
\centering
\renewcommand{\arraystretch}{1.2}
\begin{tabular}{@{}cccccc@{}}
\toprule
Category & Teacher Model & Teacher & Surrogate & Gap & Fidelity\\
\midrule
$-$ & \textsc{Tide} (ours) & 0.0081 & \textbf{0.0129} & +\textbf{0.0048} & 0.0141\\
\midrule
\multirow{2}{*}{\shortstack{Hybrid}}
& CNN-BiGRU & \textbf{0.0080} & 0.0142 & +0.0062 & 0.0143\\
& CNN-BiLSTM & 0.0085 & 0.0141 & +0.0056 & 0.0152\\
\midrule
\multirow{2}{*}{\shortstack{Variant}}
& \(C_1+C_2\) & 0.0356 & 0.0428 & +0.0072 & \textbf{0.0017}\\
& \(C_1+C_3\) & 0.0136 & 0.0188 & +0.0052 & 0.0199\\
\bottomrule
\end{tabular}
\end{table}

\subsubsection{Symbolic Surrogate}
{\color{black}To assess the approximation underlying both model-level interpretation and deployment, we compare surrogates distilled from different teachers. SoH estimation error measures agreement with the ground truth, whereas fidelity error measures agreement with teacher predictions. This distinction matters because reproducing a teacher closely does not by itself ensure accurate SoH estimation.}
Seen from Table \ref{tab:surrogate}, the \textsc{Tide} backbone is the most suitable teacher for symbolic distillation, achieving the lowest surrogate RMSE of 0.0129. Although hybrid models achieve accuracies comparable to \textsc{Tide}, their distilled symbolic equations exhibit higher RMSE compared with the surrogate of \textsc{Tide}, where the error gap from teacher to student widens by 29\% and 17\%, respectively. This indicates that high teacher accuracy alone does not guarantee an accurate symbolic surrogate. This is likely because the latent representations learned by comparison models that are neural network based are highly entangled, making them difficult to approximate using compact symbolic expressions. Instead, the structured \textsc{Tide} backbone produces representations that are better suited to symbolic approximation. Among the \textsc{Tide} variants, the monotone variant $C_1+C_2$ can be well reproduced by the symbolic surrogate with the lowest fidelity error; however, its estimation capability is limited and the surrogate error for SoH estimation is high. The contextual variant $C_1+C_3$ is considerably more difficult to approximate symbolically, with a fidelity error approximately one order of magnitude higher than that of the monotone variant. These results demonstrate that a well-designed teacher backbone is essential for obtaining an accurate and compact symbolic surrogate.
The final selected symbolic surrogate is derived from the \textsc{Tide} backbone as,
\begin{equation}
\begin{aligned}
\hat{S}
&=
\sqrt{\mathcal{G}(s_v,x_{ir},\bar{r})+\sqrt{\mathcal{A}(i,\bar{v})}},\\
\mathcal{A}(i,\bar{v})
&=
0.9006 - 0.0739(c+\bar{v}),\\
\mathcal{G}(s_v,x_{ir},\bar{r})
&=
s_v(-0.00794x_{ir}-0.02017\bar{r}-0.04988),
\end{aligned}
\label{eq:surrogate}
\end{equation}
besides the cycle index $i$ and internal resistance $r$, $\bar{v}$ denotes the mean voltage, $s_v$ the voltage variability, $\bar{r}$ the rolling internal-resistance mean, and $x_{ir}$ the cycle-resistance interaction. The surrogate in Eq. \eqref{eq:surrogate} serves as a compact representation that summarizes the behavior of the \textsc{Tide} backbone. It can be understood through three complementary symbolic terms. The aging term $\mathcal{A}(i,\bar{v})$ captures the global degradation trend from cumulative cycling and voltage context, while the interaction term $\mathcal{G}(s_v,x_{ir},\bar{r})$ summarizes contextual corrections arising from voltage variability, cycle-resistance coupling, and rolling internal resistance. Finally, the nested square-root provides a compact nonlinear transformation that integrates these effects into the final SoH estimate while preserving the overall behavior of the teacher model. {\color{black}Together, these terms provide a model-level account of the learned estimation behavior in a form that can also be evaluated directly at deployment. The coefficients are tied to the fitted preprocessing used during symbolic search, so the corresponding feature transformations are part of this inference procedure. Note that the expression represents the learned estimator within the evaluated data domain and should not be interpreted as a general electrochemical law.}

\begin{figure*}[!t]
\centering
\def \htmp{0.65\linewidth}
\def \ctmp{0.49\columnwidth}
\def \hstmp{0.1em}
\begin{subfigure}[t]{\ctmp}
    \centering
    \includegraphics[height=\htmp]{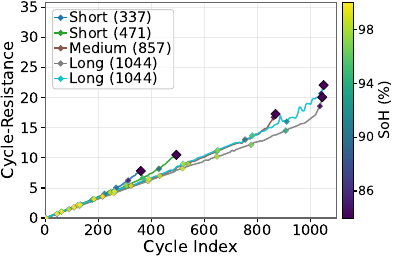}
    \caption{Cycle-Aging Fingerprint}
    \label{fig:insights-cr-c}
\end{subfigure}
\hspace{\hstmp}
\begin{subfigure}[t]{\ctmp}
    \centering
    \includegraphics[height=\htmp]{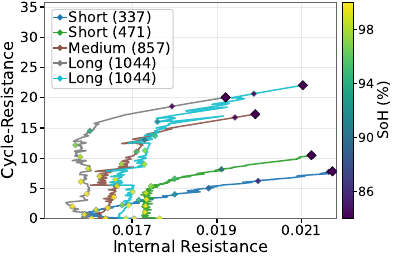}
    \caption{Cycle-Resistance Fingerprint}
    \label{fig:insights-cr-r}
\end{subfigure}
\hspace{\hstmp}
\begin{subfigure}[t]{\ctmp}
    \centering
    \includegraphics[height=\htmp]{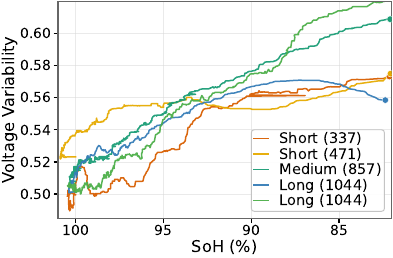}
    \caption{Voltage-Variability}
    \label{fig:insights-voltage}
\end{subfigure}
\hspace{\hstmp}
\begin{subfigure}[t]{\ctmp}
    \centering
    \includegraphics[height=\htmp]{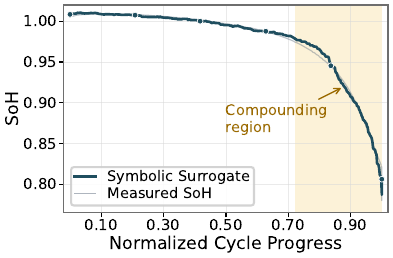}
    \caption{Late-Life Compounding}
    \label{fig:insights-late}
\end{subfigure}
\caption{Representative physical insights revealed by the symbolic surrogate. Fig. \ref{fig:insights-cr-c}: Cycle-aging patterns across batteries with different lifespans. Fig. \ref{fig:insights-cr-r}: Relationship between internal resistance and the cycle–resistance interaction. Fig. \ref{fig:insights-voltage}: Voltage-variability trajectories throughout battery degradation. Fig. \ref{fig:insights-late}: Transition from gradual aging to accelerated late-life degradation.}
\label{fig:insights}
\end{figure*}

{\color{black}
Figs.~\ref{fig:insights-cr-c}--\ref{fig:insights-voltage} relate the
symbolic expression to observed feature trajectories. The interaction
$x_{ir}=i\,r_i$ is plotted against cycle index and resistance in
Figs.~\ref{fig:insights-cr-c} and~\ref{fig:insights-cr-r}, respectively.
Fig.~\ref{fig:insights-voltage} relates voltage variability to measured
SoH. These trajectories contextualize the expression's inputs rather
than measure individual symbolic-term contributions.
Batteries are grouped by observed lifespan using the 33rd and 67th
percentiles. Two short-life, one medium-life, and two long-life examples
are selected, ranked within each group by voltage-variability span and
then record count. The same examples are used across all three panels.
Cycle--resistance trajectories are interpolated, and voltage variability
is smoothed using a centered rolling median. These display operations,
lifespan grouping, and measured-SoH annotations support retrospective
interpretation and are not additional deployment inputs.
}

 As shown in Fig. \ref{fig:insights-cr-c}, the cycle-aging fingerprint reveals distinct aging trajectories for batteries with different lifespans. Batteries with shorter lifespans accumulate aging stress more rapidly throughout the degradation process, whereas longer-life batteries exhibit more gradual increases before entering accelerated degradation near the EoL. Fig. \ref{fig:insights-cr-r} further illustrate how the cycle-resistance interaction evolves across internal resistances. Batteries with shorter lifespans enter higher interaction-stress regions earlier, whereas longer-life batteries follow more gradual trajectories, indicating that the coupled effect of cycle accumulation and resistance growth provides a interpretable aging fingerprint. 
Fig. \ref{fig:insights-voltage} further reveals that voltage variability exhibits distinct evolution patterns. Compared with shorter-life batteries, longer-life batteries sustain higher levels of voltage variability over a broader SoH range while maintaining gradually changed trajectories during the early aging stage. This suggests that voltage variability captures battery-specific operational characteristics beyond conventional indicators such as the raw voltage readings.
Finally, Fig. \ref{fig:insights-late} reveals that the symbolic surrogate quantitatively characterizes the transition from gradual to accelerated degradation. The highlighted late-life compounding region corresponds to the near EoL stage degradation, beginning around the knee point of the degradation curve to emphasize the transition from gradual to accelerated SoH decline. Before this region, the predicted SoH decreases by around $2.5\%$ only. Once the battery enters this compounding region, the SoH drops significantly. This substantial SoH loss indicates that cumulative aging mechanisms progressively accelerate battery deterioration, and the symbolic surrogate can provide a mathematical characterization. 

\balance
\section{Conclusion}
\label{sec:conclusion}
This paper presented \textsc{Tide}, a trustworthy and interpretable framework for battery SoH estimation. It integrates degradation knowledge with operational sensing features through a three-component backbone comprising a knowledge prior, a monotone residual and a contextual component. Compared with representative baselines, \textsc{Tide} improves estimation accuracy by an average of 19.7\%, achieves the highest $R^2$ of 0.964. It also produces the lowest monotonicity violations for cycle index and internal resistance, whereas the competitive hybrid baselines exhibit violation rates above 30\%. {\color{black}The \textsc{Tide} backbone provides component-level interpretations, and symbolic distillation yields a compact surrogate that summarizes the complete estimation process and characterizes patterns such as cycle-resistance coupling and late-stage degradation. The model representation also offers a route to deployment through direct expression evaluation, with approximation accuracy and computational cost determining its suitability for a given battery system.} Overall, \textsc{Tide} jointly advances accuracy, trustworthiness, and interpretability for practical battery analytics.

{\color{black}In the future, we would like to explore alternative battery degradation priors within the proposed framework, evaluate the symbolic surrogate on embedded BMS hardware under noisy and incomplete sensing, and adapt the framework to industrial equipment health monitoring using domain-specific knowledge and measurements.}

\balance

\bibliographystyle{IEEEtran}
\bibliography{reference}

\end{document}